\documentclass[conference]{IEEEtran}
\usepackage{cite}
\usepackage{amsmath,amssymb,amsfonts}
\usepackage{algorithm}
\usepackage{algorithmic}
\usepackage{graphicx}
\usepackage{textcomp}
\usepackage{xcolor}
\def\BibTeX{{\rm B\kern-.05em{\sc i\kern-.025em b}\kern-.08em
    T\kern-.1667em\lower.7ex\hbox{E}\kern-.125emX}}

\usepackage[hyphens]{url}
\usepackage{hyperref}
\usepackage[hyphenbreaks]{breakurl}

\usepackage[switch]{lineno}

\usepackage{amsthm}

\newtheorem{theorem}{Theorem}
\newtheorem{definition}{Definition}
\newtheorem{assumption}{Assumption}

\usepackage{listings}
\floatstyle{ruled}
\newfloat{listing}{tb}{lst}{}
\floatname{listing}{Listing}

\newcommand{\refassumption}[1]{Assumption~\ref{#1}}
\newcommand{\refdefinition}[1]{Definition~\ref{#1}}

\newcommand{\reffigure}[1]{\figurename~\ref{#1}}
\newcommand{\refsection}[1]{Section~\ref{#1}}
\newcommand{\refsubsection}[1]{Subsection~\ref{#1}}

\newcommand{\reftheorem}[1]{Theorem~\ref{#1}}

\begin{document}

\title{The Ludii Game Description Language is Universal}


\author{\IEEEauthorblockN{Dennis J.N.J. Soemers,\IEEEauthorrefmark{1} {\'E}ric Piette,\IEEEauthorrefmark{2} Matthew Stephenson,\IEEEauthorrefmark{3} and Cameron Browne\IEEEauthorrefmark{4}}
\IEEEauthorblockA{\IEEEauthorrefmark{1}\textit{Department of Advanced Computing Sciences, Maastricht University}\\ Email: \texttt{dennis.soemers@maastrichtuniversity.nl}}
\IEEEauthorblockA{\IEEEauthorrefmark{2}\textit{ICTEAM, Universit{\'e} catholique de Louvain}\\
Email: \texttt{eric.piette@uclouvain.be}}
\IEEEauthorblockA{\IEEEauthorrefmark{3}\textit{College of Science and Engineering, Flinders University}\\
Email: \texttt{matthew.stephenson@flinders.edu.au}}
\IEEEauthorblockA{\IEEEauthorrefmark{4}Email: \texttt{cambolbro@gmail.com}}
}

\maketitle

\begin{abstract}
There are several different game description languages (GDLs), each intended to allow wide ranges of arbitrary games (i.e., general games) to be described in a single higher-level language than general-purpose programming languages. Games described in such formats can subsequently be presented as challenges for automated general game playing agents, which are expected to be capable of playing any arbitrary game described in such a language without prior knowledge about the games to be played. The language used by the Ludii general game system was previously shown to be capable of representing equivalent games for any arbitrary, finite, deterministic, fully observable extensive-form game. In this paper, we prove its universality by extending this to include finite non-deterministic and imperfect-information games.
\end{abstract}


\section{Introduction}

General Game Playing (GGP) is a subfield of Artificial Intelligence (AI) research, in which the challenge is to develop agents that can successfully play arbitrary games without human intervention or prior knowledge of exactly which games are to be played \cite{Pitrat_1968_GGP}. Implementing such an agent in practice typically requires the use of a Game Description Language (GDL); a standardised format such that the rules of any game can be provided to an agent without having to implement it directly in a general-purpose programming language. 

The GDL that popularised GGP research \cite{Genesereth_2005_GGP,Love_2008_GDL} originated primarily from Stanford; we refer to it as S-GDL in this paper. Other systems with GDLs include Regular Boardgames (RBG) \cite{Kowalski_2019_Regular} and Ludii \cite{Browne_2020_Practical,Piette_2020_Ludii} for general games, as well as GVGAI \cite{Schaul_2014_Extensible,Perez_2019_GVGAI} for video games. Aside from facilitating GGP research, the use of domain-specific languages has also been proposed for the ease with which they enable the implementation of custom, targeted testbeds \cite{Samvelyan_2021_MiniHack}.

S-GDL is a relatively low-level logic-based GDL. After the introduction of an extension to support randomness and imperfect information \cite{Thielscher_2010_GDLII}, it was proven that S-GDL is \textit{universal} \cite{Thielscher_2011_Universal}; any arbitrary finite extensive-form \cite{Rasmusen_2007_Games} game can be faithfully represented in a legal S-GDL description. For the GDL of RBG, this was only proven for the subset of fully-observable, deterministic games \cite{Kowalski_2019_Regular}. Similarly, Ludii's GDL (L-GDL) was previously only proven to be capable of representing any finite, deterministic, perfect-information, alternating-move game, although it did already include basic support for stochasticity and hidden information (without a proof of universality) \cite{Piette_2020_Ludii}.

For S-GDL, the proof of its universality \cite{Thielscher_2011_Universal} essentially consists of encoding the entire game tree of any arbitrary finite extensive-form game in logic statements. L-GDL is a comparatively higher-level language that primarily consists of many keywords that game designers and players can readily understand as common game terms, such as \texttt{board}, \texttt{piece}, \texttt{slide}, \texttt{hop}, and so on. By design, it is intended to be easier to read, understand and use for game designers \cite{Browne_2016_Class}, with less of a focus on including the low-level language elements that would enable the exhaustive enumeration of all states of an extensive-form game tree. It has a relatively tightly-enforced structure, with many enforced restrictions due to strong typing. In comparison to the lower-level S-GDL with a relatively flat structure that makes it straightforward to exhaustively enumerate a complete game tree, this makes it non-trivial to prove a similar level of generality for L-GDL. Nevertheless, in this paper we are able to prove the universality of L-GDL by demonstrating that it can represent the same class of games as proven by Thielscher~\cite{Thielscher_2011_Universal} for S-GDL, including games with randomness and hidden information. This provides a theoretical argument that L-GDL is a suitable, sufficiently general and powerful description language for problems for AI research.

The remainder of this paper is structured as follows. \refsection{Sec:Background} provides the necessary background information on extensive-form games and the L-GDL game description language. Next, \refsection{Sec:EFG_to_LGDL} proposes a detailed procedure that, for any finite extensive-form game $\mathcal{G}$, creates a matching L-GDL game description for a Ludii game $\mathcal{G}^L$. \refsection{Sec:ProofEquivalence} formally states a theorem of equivalence for $\mathcal{G}$ and $\mathcal{G}^L$, and proves the theorem. In \refsection{Sec:Discussion}, we provide a brief discussion of two related topics that may be considered of interest around the main theorem. Finally, \refsection{Sec:Conclusion} concludes the paper.

\section{Background} \label{Sec:Background}

In this section, we provide background information on the standard, universal formalism of extensive-form games, as well as L-GDL. 

\subsection{Extensive-Form Games}

Extensive-form games \cite{Rasmusen_2007_Games} are a standard, general formalisation of games in the broad, mathematical sense of the word (i.e., including many decision-making problems that would not generally be viewed by most humans as ``fun'' games). The formal definition is as follows:

\begin{definition} \label{Def:ExtensiveFormGame}
An extensive-form game $\mathcal{G}$ is specified by a tuple $\mathcal{G} = \langle \mathcal{P}, \mathcal{T}, \mathcal{U}, \iota, \mathcal{D}, \mathcal{I} \rangle$, where:

\begin{itemize}
    \item $\mathcal{P} = \{ 1, 2, \dots, k, \eta \}$ is a finite set of $k \geq 1$ players, and a ``nature'' player $\eta$ to model stochastic events.
    \item $\mathcal{T}$ is a finite tree, where every node represents a single game state $s \in \mathcal{S}$. The full set of states $\mathcal{S} = \mathcal{S}_{inn} \cup \mathcal{S}_{ter}$ may be partitioned into a subset of non-terminal states (inner nodes) $\mathcal{S}_{inn}$ and a subset of terminal states (leaf nodes) $\mathcal{S}_{ter}$, such that $\mathcal{S}_{inn} \cap \mathcal{S}_{ter} = \varnothing$. Every branch of the tree represents a possible transition between states.
    \item $\mathcal{U} : \mathcal{S}_{ter} \mapsto \mathbb{R}^k$ is a payoffs function, such that $\mathcal{U}(s)$ denotes a vector of $k$ real-valued payoffs (for the $k$ players) for any terminal game state $s \in \mathcal{S}_{ter}$.
    \item $\iota : \mathcal{S}_{inn} \mapsto \mathcal{P}$ is a function such that, for any non-terminal game state $s \in S_{inn}$, $\iota(s)$ gives the player to play in that state. Whenever $\iota(s) \neq \eta$ (i.e., whenever we are not in a chance node), the player gets to choose which branch to follow down the tree (it is not permitted to go back up to the parent node).
    \item $\mathcal{D} : \{ (s, s') \mid \iota(s) = \eta, s \in \mathcal{S}_{inn}, s' \in \mathcal{S} \} \mapsto \mathbb{R}$ gives, for any non-terminal state $s$ controlled by the nature player $\eta$, a probability $0 \leq \mathcal{D}(s, s') \leq 1$ that the nature player ``selects'' $s'$ as the successor. Note that this must yield proper probability distributions over successors, i.e. $\forall s \in \{ s \mid \iota(s) = \eta, s \in \mathcal{S}_{inn} \}: \sum_{s' \in \mathcal{S}} \mathcal{D}(s, s') = 1$.
    \item $\mathcal{I}: \{ (p, s) \mid p \in \mathcal{P} \setminus \{ \eta \}, s \in \mathcal{S} \} \mapsto \mathbb{P}(\mathcal{S})$, where $\mathbb{P}(\mathcal{S})$ denotes the powerset of $\mathcal{S}$, gives the information set $\mathcal{I}(p, s)$ of player $p$ for state $s$ (i.e., the set of states that are indistinguishable from each other from the perspective of player $p$ when the true state is $s$).
\end{itemize}

\end{definition}

In this paper, we focus on finite extensive-form games $\mathcal{G}$, where $\mathcal{T}$ is of a finite size. Furthermore, we focus on sequential-move games, since the function $\iota$ gives only a single player to move per game state $s$. In theory, this is without loss of generality, since any simultaneous-move game can be equivalently modelled as an sequential-move game in which the effects of moves are delayed until every active player in a turn has selected their move, and moves within the same turn are hidden information for all other players \cite{Watson_2013_Strategy}. In practice, Ludii does contain additional support for modelling simultaneous-move games, but for our theoretical analysis we do not need this.

\subsection{L-GDL}

The basic structure of an L-GDL game description is depicted in \reffigure{Fig:StructureLudiiGameDescription}. It is defined by a grammar \cite{Browne_2020_LLR}, automatically derived from Ludii's source code \cite{Browne_2016_Class}, which specifies which keywords (also referred to as \textit{ludemes}) and types of data (strings, integers, real numbers, and so on) can or cannot be used depending on the context. As shown by \reffigure{Fig:StructureLudiiGameDescription}, a game description file is expected to describe exactly one game, which has three top-level entries:
\begin{enumerate}
    \item \texttt{players}: describes basic data about the players (e.g., how many players the game is played by).
    \item \texttt{equipment}: describes aspects such as any board(s) or graph(s) the game is played on, types of pieces or dice used in the game, and so on.
    \item \texttt{rules}: describes rules used to (i) \textbf{start} the game (generate initial game state, e.g. by placing initial pieces on a board), (ii) \textbf{play} the game (generate lists of legal moves), and (iii) \textbf{end} the game (evaluate whether a state is terminal and determine the outcomes for the players).
\end{enumerate}
Some of these aspects must be specified (such as the play rules), whereas others may be omitted if unnecessary (e.g., start rules are unnecessary in games that start with an empty board) or if they have a suitable default value (e.g., Ludii assumes a default number of players of $2$ if left unspecified). L-GDL includes a relatively large set of ludemes, many of which encapsulate relatively high-level concepts in keywords that game designers can easily understand and use to write and read game descriptions. Piette et al.~\cite{Piette_2020_Ludii} provide more detailed information on the Ludii system, and Browne et al.~\cite{Browne_2020_LLR} provide a complete, detailed language reference for L-GDL.

A full example description for the game of Tic-Tac-Toe is presented in \reffigure{Fig:TicTacToeDescription}. In this example, the equipment used to play the game is defined as a square board of size $3$ (by default using a tiling of square cells), a ``Disc'' piece type used by player 1, and a ``Cross'' piece type used by player 2. The subtree of ludemes \texttt{(move Add (to (sites Empty)))} describes that the set of legal moves consists of moves that add a piece to any site in the set of empty sites. The subtree \texttt{(if (is Line 3) (result Mover Win))} describes the end condition of this game, which is that the current mover wins if they complete a contiguous line consisting of 3 of their pieces.

\section{From Extensive-Form Games to L-GDL} \label{Sec:EFG_to_LGDL}

Given any arbitrary finite, extensive-form game $\mathcal{G} = \langle \mathcal{P}, \mathcal{T}, \mathcal{U}, \iota, \mathcal{D}, \mathcal{I} \rangle$ as defined in \refdefinition{Def:ExtensiveFormGame}, we describe how a corresponding Ludii game $\mathcal{G}^L$ can be modelled in L-GDL. In \refsection{Sec:ProofEquivalence}, we formally state and prove the theorem that $\mathcal{G}$ and $\mathcal{G}^L$ form equivalent game trees with one-to-one correspondences between the set of all possible trajectories in $\mathcal{G}$ and the set of all possible trajectories in $\mathcal{G}^L$. For simplicity, and without loss of generality, we make several assumptions about $\mathcal{G}$:

\begin{assumption} \label{Assumption:InitialGameState}
$\mathcal{G}$ has a unique initial game state $s_0$ as root node of its game tree.
\end{assumption}
This assumption is without loss of generality because a game with multiple distinct possibilities for the initial game state can be equivalently modelled as a game with a single chance node as root, with appropriate probabilities assigned for all the intended ``real'' initial game states.
\begin{assumption} \label{Assumption:Player1First}
If the root node of $\mathcal{G}$ is not a chance node, the player labelled as $1$ will be the first player to make a move.
\end{assumption}
This assumption is without loss of generality because there is otherwise no particular meaning to the labels that are assigned to players.

\begin{figure}[t]
\centering
\begin{lstlisting}
(<@\textbf{game}@> "Game Name"
  (<@\textbf{players}@> <@$\dots$@>)
  (<@\textbf{equipment}@> {
    <@$\dots$@>
  })
  (<@\textbf{rules}@>
    (<@\textbf{start}@> <@$\dots$@>)
    (<@\textbf{play}@> <@$\dots$@>)
    (<@\textbf{end}@> <@$\dots$@>)
  )
)
\end{lstlisting}
\caption{Basic structure of an L-GDL game description for Ludii. Note that curly braces are used for arrays in L-GDL.}
\label{Fig:StructureLudiiGameDescription}
\end{figure}

\begin{figure}[t]
\centering
\begin{lstlisting}
(<@\textbf{game}@> "Tic-Tac-Toe"
  (<@\textbf{players}@> <@$2$@>)
  (<@\textbf{equipment}@> {
    <@(board (square 3))@>
    <@(piece "Disc" P1)@>
    <@(piece "Cross" P2)@>
  })
  (<@\textbf{rules}@>
    (<@\textbf{play}@> <@(move Add (to (sites Empty)))@>)
    (<@\textbf{end}@> <@(if (is Line 3) (result Mover Win))@>)
  )
)
\end{lstlisting}
\caption{Full L-GDL description for the game of Tic-Tac-Toe.}
\label{Fig:TicTacToeDescription}
\end{figure}

The following subsections describe how to fill in the basic template L-GDL description from \reffigure{Fig:StructureLudiiGameDescription} to construct such a Ludii game $\mathcal{G}^L$. The intuition behind our approach is similar to that of the proof by Piette \textit{et al.}~\cite{Piette_2020_Ludii} (which was restricted to deterministic, perfect-information settings) in the sense that we explicitly enumerate the entire game tree of $\mathcal{G}$ as a graph that the players play on by moving stones along a path from the root to any leaf. The most significant change is that, to support imperfect-information settings, we now use multiple ``copies'' of such a graph, with sets of possibly more than one stone per player moving down each player's respective tree to track the information sets (rather than individual states) that players navigate between. 
As an example, a full game description file for the Monty Hall problem, described as explained in the following five subsections, is provided in the Ludii github repository.\footnote{\url{https://github.com/Ludeme/Ludii/blob/master/Common/res/lud/test/dennis/MontyHallProblemExtensiveForm.lud}}
This problem involves partial observability and stochasticity.

\subsection{Defining the Players} \label{Subsec:DefiningPlayers}

For a $k$-player extensive-form game $\mathcal{G}$ with players $\mathcal{P} = \{1, 2, \dots, k, \eta\}$, the set of players in Ludii can simply be defined as \texttt{(players k)}. It is not necessary to explicitly define the nature player in Ludii. The player labelled as player $1$ in Ludii will, by default, be the first player to make a move, matching \refassumption{Assumption:Player1First}.



\subsection{Defining the Equipment} \label{Subsec:DefiningEquipment}

Firstly, we define a neutral \texttt{Marker0} piece type---which we use to keep track of the true game state that we are in during any given trajectory of play---as well as one \texttt{MarkerP} piece type for every player $1 \leq P \leq k$---which are used to reveal the correct information set to each player. This equipment is defined in the \texttt{equipment(\{$\dots$\})} section of the game description using \texttt{(piece "Marker" Neutral)} and \texttt{(piece "Marker" Each)}.

Secondly, we construct the game board by defining a graph that contains $(k + 1) \times \vert \mathcal{S} \vert$ vertices. These may be thought of as representing $(k + 1)$ copies of the game tree $\mathcal{T}$ in $\mathcal{G}$, with $\vert \mathcal{S} \vert$ vertices per copy, although it is not necessary to also include the connectivity structure (i.e., the edges of the tree) in this graph. Hence, the game board consists of one large graph, which contains separate graph representations of the full game tree for each player (including the neutral player). Such a graph can be constructed manually using \texttt{(graph vertices:\{$\dots$\})}. Let $i$ denote the unique index of a state $s_i \in \mathcal{S}$. Then, in the graph for player $p$ (assume $p = 0$ for the nature player), the index of the node that corresponds to $s_i$ is given by $p \times \vert \mathcal{S} \vert + i$. Without loss of generality, we assume that the index of the initial game state is $0$.

Thirdly, for every state $s_i \in \mathcal{S}$ and every player $1 \leq p \leq k$, we define a region in the equipment that contains all the indices of the vertices corresponding to states that are in the information set of $p$ when the true state is $s_i$. More formally, for all $0 \leq i < \vert \mathcal{S} \vert$ and all $1 \leq p \leq k$, we define a region named \texttt{"InformationSet\_i\_p"} containing all the indices $p \times \vert \mathcal{S} \vert + j$ for all $j \in \{ j \mid s_j \in \mathcal{I}(p, s_i) \}$. Such a region can be defined in a game description using \texttt{(regions "InformationSet\_i\_p" \{$\dots$\})}. Whenever the true state is $s_i$, this region allows us to easily access all the vertices corresponding to the complete information set for any given player $p$.

Fourthly, for every player $1 \leq p \leq k$, we define a region in the equipment named \texttt{"Subgraph\_p"} that contains all the indices of the vertices in that player's respective subgraph, i.e. all indices in $\{j \mid p \times \vert\mathcal{S}\vert \leq j < (p + 1) \times \vert\mathcal{S}\vert \}$. Such a region can be defined in a game description using \texttt{(regions "Subgraph\_p" \{$\dots$\})}. We similarly define a region \texttt{"Subgraph\_0"} for the first subgraph.

Note that the definitions of piece types, graphs, and regions as detailed above do not yet have many semantics associated with them. These statements largely serve to declare the existence of various types of data, such that they may be referenced (by their names) and used in the definitions of rules as described in the subsequent subsections. \reffigure{Fig:ProofEquipment} provides a template for an equipment definition following the steps that were just listed.

\begin{figure}[t]
\centering
\begin{lstlisting}
<@$\dots$@>
(<@\textbf{equipment}@> {
  (piece "Marker" Neutral)
  (piece "Marker" Each)
  (board
    (graph
      vertices:{
        // Vertices for tracking game state
        {<@$x_1 \; y_1$@>} <@$\dots$@> {<@$x_{\vert \mathcal{S} \vert} \; y_{\vert \mathcal{S} \vert}$@>}
        // Vertices for tracking first infoset
        {<@$x_1 \; y_1$@>} <@$\dots$@> {<@$x_{\vert \mathcal{S} \vert} \; y_{\vert \mathcal{S} \vert}$@>}
        <@$\dots$@>
        // Vertices for tracking <@$k^{th}$@> infoset
        {<@$x_1 \; y_1$@>} <@$\dots$@> {<@$x_{\vert \mathcal{S} \vert} \; y_{\vert \mathcal{S} \vert}$@>}
      }
    )
    use:Vertex
  )
  // For every state and every player, an infoset 
  // listing all the possible states
  (regions "InformationSet_0_1" {<@$\dots$@>})
  <@$\dots$@>
  (regions "InformationSet_<@$\vert \mathcal{S} \vert$@>_1" {<@$\dots$@>})
  (regions "InformationSet_0_2" {<@$\dots$@>})
  <@$\dots$@>
  (regions "InformationSet_<@$\vert \mathcal{S} \vert$@>_<@$k$@>" {<@$\dots$@>})
  // Each player has a copy of the tree
  (regions "Subgraph_0" {0..<<@$\vert \mathcal{S} \vert - 1$@>>})
  (regions "Subgraph_1" {<@$\vert \mathcal{S} \vert$@>..<<@$2 \times \vert \mathcal{S} \vert - 1$@>>})
  <@$\dots$@>
  (regions "Subgraph_<@$k$@>" {<<@$k \times \vert \mathcal{S} \vert$@>>..<<@$(k + 1) \times \vert \mathcal{S} \vert - 1$@>>})
})
<@$\dots$@>
\end{lstlisting}
\caption{Template for the equipment definition of a Ludii game $\mathcal{G}^L$, modelling an equivalent extensive-form game $\mathcal{G}$ with $\vert \mathcal{S} \vert$ different states and $k$ players. The expressions angled brackets are used for generality, but would be replaced by the concrete result of the expression in any single concrete game description. The values used for $x$- and $y$-coordinates only affect display in Ludii's graphical user interface, and are irrelevant in terms of semantics.}
\label{Fig:ProofEquipment}
\end{figure}

\subsection{Defining the Start Rules} \label{Subsec:DefiningStartRules}

Due to \refassumption{Assumption:InitialGameState}, we know that every player's information set for the initial game state contains only $s_0$; $\forall_p \mathcal{I}(p, s_0) = \{ s_0 \}$. Hence, we start the game by placing a marker for each player (including a neutral marker for the nature player) on the vertex that represents the initial game state in each player's respective subgraph in the board. For any player $p$, the index of this vertex is given by $p \times \vert \mathcal{S} \vert$, assuming $p = 0$ for the nature player. Presence or absence of markers on any site in a subgraph corresponding to a player $p$ must be hidden from all other players $p' \neq p$, to avoid leaking information that those other players should not have access to. 

Start rules that accomplish this setup for the initial game state are provided in \reffigure{Fig:ProofStartRules}. Each of the \texttt{(\textbf{place} "Marker$k$" <$x$>)} lines places a Marker for player $k$ on site $x$, marking the information set that player $k$ believes the game is in. In the initial state, every player's information set contains only a single state, thanks to \refassumption{Assumption:InitialGameState}. The marker for $k=0$ does not correspond to any particular player, but is used to mark the true game state. The ludeme \texttt{(\textbf{set} Hidden (sites $x$) to:$y$)} states that all sites in a region $x$ are set to be hidden (i.e., unobservable) to player(s) $y$. The combination of all such lines in the figure ensures that the first copy of the game tree (with index $0$) is hidden to all players, and every other copy is only observable by its respective player.

More formally, for any Ludii game $\mathcal{G}^L$ with these start rules, the following statements hold in the initial game state:
\begin{itemize}
    \item There is a piece of type \texttt{Marker0} (not owned by any player) on the vertex with index 0.
    \item For every player index $P \in [1, \dots, k]$, there is a piece of type \texttt{MarkerP} (owned by player $P$) on the vertex with index $P \times \vert \mathcal{S} \vert$.
    \item No pieces are placed other than those mentioned above.
    \item All vertices and their contents in the region named \texttt{"Subgraph\_0"} are set to be hidden to all players. This means that no player can observe that vertex 0 contains a \texttt{Marker0} piece. Similarly, none of the players can observe that all other vertices in this region are empty.
    \item For every player index $P \in [1, \dots, k]$, and every other player index $P' \in [1, \dots, k]$, with $P' \neq P$, all vertices in any region named \texttt{"Subgraph\_P"} are hidden from $P'$. This means that every player $P$ can only observe vertices (and markers placed on them) in the region named \texttt{"Subgraph\_P"}.
\end{itemize}

\begin{figure}[t]
\centering
\begin{lstlisting}
<@$\dots$@>
(<@\textbf{start}@> {
  (<@\textbf{place}@> "Marker0" <@$0$@>)
  (<@\textbf{place}@> "Marker1" <<@$1 \times \vert \mathcal{S} \vert$@>>)
  <@$\dots$@>
  (<@\textbf{place}@> "Marker<@$k$@>" <<@$k \times \vert \mathcal{S} \vert$@>>)
  (<@\textbf{set}@> Hidden 
    (sites "Subgraph_0") to:All)
  (<@\textbf{set}@> Hidden 
    (sites "Subgraph_1") to:(player 2))
  <@$\dots$@>
  (<@\textbf{set}@> Hidden 
    (sites "Subgraph_1") to:(player <@$k$@>))
  (<@\textbf{set}@> Hidden 
    (sites "Subgraph_2") to:(player 1))
  (<@\textbf{set}@> Hidden 
    (sites "Subgraph_2") to:(player 3))
  <@$\dots$@>
  (<@\textbf{set}@> Hidden 
    (sites "Subgraph_2") to:(player <@$k$@>))
  <@$\dots$@>
  (<@\textbf{set}@> Hidden 
    (sites "Subgraph_<@$k$@>") to:(player <@$k - 1$@>))
})
<@$\dots$@>
\end{lstlisting}
\caption{Start rules for a Ludii game $\mathcal{G}^L$, modelling an equivalent extensive-form game $\mathcal{G}$, with $k$ players. The expressions to compute vertex indices in angled brackets are used for generality, but would be replaced by the concrete result of the expression in any single concrete game description.}
\label{Fig:ProofStartRules}
\end{figure}

\subsection{Defining the Play Rules} \label{Subsec:DefiningPlayRules}

The play rules in Ludii define how to generate a list of legal moves for any given current game state $s_i$. In our $\mathcal{G}^L$ model, where we aim to replicate the structure of the game tree of the extensive-form game $\mathcal{G}$, we may distinguish two primary cases:
\begin{enumerate}
    \item If $s_i$ is a chance node, i.e. $\iota(s_i) = \eta$, in Ludii a regular player will be in control because Ludii does not explicitly include a nature player. Hence, we should generate only a single legal move such that the player is forced to traverse the branch that the chance player would have picked in $\mathcal{G}$. This can be accomplished by using the \texttt{(random \{$\dots$\} \{$\dots$\})} ludeme, where the first array contains a sequence of $n$ weights, and the second array contains a sequence of $n$ move-generating ludemes, for a chance node with $n$ possible branches. For example, \texttt{(random \{ $p$, $q$, $r$ \} \{ A B C \})} randomly selects one of the ludemes \texttt{A}, \texttt{B}, or \texttt{C} to generate the list of legal moves, with probabilities $\frac{p}{p + q + r}$, $\frac{q}{p + q + r}$, or $\frac{r}{p + q + r}$, respectively. The appropriate weights to use can be derived from the nonzero probabilities $\mathcal{D}(s, s')$ as specified in $\mathcal{G}$.
    \item If $s_i$ is not a chance node, i.e. $\iota(s_i) \neq \eta$, the mover $\iota(s_i)$ should have one move corresponding to every branch from $s_i$ in the game tree of the extensive-form game $\mathcal{G}$. This can be implemented using an \texttt{(or \{ $\dots$ \})} ludeme that wraps around other ludemes, each of which generates one of the legal moves.
\end{enumerate}

Without any knowledge of any general rules that may determine how legal moves are computed from a game state $s_i$ in the extensive-form game $\mathcal{G}$, it is necessary to explicitly enumerate all game states and define the play rules separately per state. One way to accomplish this is by using a chain of \texttt{(if C A B)} ludemes, where:
\begin{itemize}
    \item \texttt{C} is a condition of the form \texttt{(= (where "Marker" Neutral) i)}: this checks whether the \texttt{Marker0} piece is located on vertex $i$, and can hence be used to determine whether or not the current game state is $s_i$.
    \item \texttt{A} is a ludeme that generates the moves in the case that the condition of \texttt{C} is satisfied by the current game state (i.e., if the current game state is $s_i$).
    \item \texttt{B} is a ludeme that generates the moves if the current game state does not satisfy the condition of \texttt{C}; this can again be a ludeme of the same \texttt{(if C A B)} form.
\end{itemize}

Suppose that there is some branch in the extensive-form game tree of $\mathcal{G}$ that leads from a state $s_i$ to a state $s_j$. In the corresponding Ludii game $\mathcal{G}^L$, we require a corresponding move that has the following effects on the game state:
\begin{enumerate}
    \item It should move the \texttt{Marker0} piece, which should currently be located on the vertex with index $i$, to the vertex with index $j$. This enables us to continue tracking the true game state.
    \item For every player $1 \leq p \leq k$, any \texttt{Marker$p$} pieces currently on the board should be removed, and new \texttt{Marker$p$} pieces should be placed on all vertices in the \texttt{InformationSet\_$j$\_$p$} region. This enables us to let every player know which information set it transitioned into. 
    \item By default, Ludii reveals information about positions that become empty. Because the above effects remove some pieces from positions that should still remain hidden from many players, vertices should be appropriately set to hidden again as they were originally set in the start rules.
    \item By default, Ludii updates the index of which player is designated the mover after every move, by incrementing it or resetting it to $1$ after player $k$ made a move. If this results in a different player to move than the player $\iota(s_j)$ that should become the mover in $s_j$, we need to include an extra effect in the move that correctly sets the player to move. Note that, if $\iota(s_j) = \eta$ in $\mathcal{G}$, it does not matter which player is set to be the mover in Ludii, since we only generate one legal move anyway that whichever player is the mover will be forced to pick.
\end{enumerate}
Suppose that such a state $s_i$ has $n$ legal moves. A straightforward way to present $n$ different options to the player $\iota(s_i)$ is to allow them to select one out of any of vertex $0 \leq v < n$, and to specify appropriate consequences for each of those ``select'' moves. These consequences should correspond to the $v^{th}$ branch from $s_i$ in the game tree of $\mathcal{G}$, but otherwise do not necessarily have any particular relationship with the specific vertex $v$; selecting vertices is simply a mechanism through which the player can distinguish between $n$ different moves. 

\reffigure{Fig:ProofMoveRule} depicts the specification of a move rule for a single transition from $s_i$ to $s_j$. The \texttt{(move Select (from $n$) (then $\dots$))} ludeme defines a legal move where the player may opt to select a vertex $n$, which will lead to consequences as defined inside the \texttt{(then $\dots$)} ludeme. These consequences correspond to the four types of effects listed previously. The marker tracking the true game state (hidden to all players) is moved by \texttt{(fromTo (from $i$) (to $j$))}. The \texttt{(remove (sites Occupied by:P$k$))} ludemes each remove all markers for one player $k$ from the board, and each of the \texttt{(add (piece $k+1$) (to (sites "InformationSet\_$j$\_$k$")))} ludemes similarly places new markers to mark the new information set for a player $k$. Note that there is an offset of $+1$ due to the presence of the neutral piece in the game. As in \reffigure{Fig:ProofStartRules}, the \texttt{(\textbf{set} Hidden $\dots$)} ludemes ensure that no player can observe any information they should not be able to. Finally, the \texttt{(set NextPlayer (player $\iota$($s_j$)))} ludeme ensures that the correct player to move is set in the subsequent game state.

\begin{figure}[t]
\centering
\begin{lstlisting}
  <@$\dots$@>
  (move Select (from <@$n$@>)
    (then (and {
      (fromTo (from <@$i$@>) (to <@$j$@>))
      (remove (sites Occupied by:P1))
      (remove (sites Occupied by:P2))
      <@$\dots$@>
      (remove (sites Occupied by:P<@$k$@>))
      (add (piece 2) 
        (to (sites "InformationSet_<@$j$@>_1")))
      (add (piece 3) 
        (to (sites "InformationSet_<@$j$@>_2")))
      <@$\dots$@>
      (add (piece <@$k+1$@>) 
        (to (sites "InformationSet_<@$j$@>_<@$k$@>")))
      (<@\textbf{set}@> Hidden 
        (sites "Subgraph_0") to:All)
      (<@\textbf{set}@> Hidden 
        (sites "Subgraph_1") to:(player 2))
      <@$\dots$@>
      (<@\textbf{set}@> Hidden 
        (sites "Subgraph_1") to:(player <@$k$@>))
      (<@\textbf{set}@> Hidden 
        (sites "Subgraph_2") to:(player 1))
      (<@\textbf{set}@> Hidden 
        (sites "Subgraph_2") to:(player 3))
      <@$\dots$@>
      (<@\textbf{set}@> Hidden 
        (sites "Subgraph_2") to:(player <@$k$@>))
      <@$\dots$@>
      (<@\textbf{set}@> Hidden 
        (sites "Subgraph_<@$k$@>") to:(player <@$k - 1$@>))
      (set NextPlayer (player <@$\iota(s_j)$@>))
    }))
  )
  <@$\dots$@>
\end{lstlisting}
\caption{Ludeme generating a move corresponding to the $n^{th}$ branch from a state $s_i$, leading to a state $s_j$, in the game tree of an extensive-form game $\mathcal{G}$ with $k$ players. The player that should be the mover in the next state $s_j$ is denoted by $\iota(s_j)$---except we replace it by any arbitrary integer in $[1, k]$ if $\iota(s_j) = \eta$.}
\label{Fig:ProofMoveRule}
\end{figure}

\subsection{Defining the End Rules} \label{Subsec:DefiningEndRules}

For each of the terminal game states $s_t \in \mathcal{S}_{ter}$ in $\mathcal{G}$, we can define a separate end rule in $\mathcal{G}^L$ that checks whether that specific state has been reached by tracking the position of the \texttt{Marker0} piece, and assigns a vector of payoffs to the $k$ players as given by $\mathcal{U}(s_t)$ using the \texttt{payoffs} ludeme. \reffigure{Fig:ProofEndRules} provides an example of such end rules for an example game for $k=3$ players with two terminal game states.

\begin{figure}[t]
\centering
\begin{lstlisting}
(end {
  (if (= (where "Marker" Neutral) 88) 
    (payoffs {
      (payoff P1 -1) 
      (payoff P2 0.5) 
      (payoff P3 1)
    })
  )
  (if (= (where "Marker" Neutral) 2077) 
    (payoffs {
      (payoff P1 10) 
      (payoff P2 12) 
      (payoff P3 2020)
    })
  )
})
\end{lstlisting}
\caption{Example end rules for an example game with $k=3$ players, where states $s_{88}$ and $s_{2077}$ are terminal states, with payoff vectors of $[-1, 0.5, 1]$ and $[10, 12, 2020]$, respectively.}
\label{Fig:ProofEndRules}
\end{figure}

In the end rules, a ludeme of the form \texttt{(if (A) (payoffs $\dots$))} ensures that, as soon as a game state is reached where the condition \texttt{(A)} holds, the game terminates and payoffs are assigned to all players as per the \texttt{(payoffs $\dots$)} ludeme. A condition of the form \texttt{(= (where "Marker" Neutral) $x$)}, as used in \reffigure{Fig:ProofEndRules}, is true in any game state where there is a piece of type \texttt{Marker0} on the vertex with index $x$. Given the construction of the start and play rules described previously, this means that we check whether we are in the state meant to represent state $s_x$ from the original extensive-form game $\mathcal{G}$. A ludeme of the form \texttt{(payoffs {(payoff P1 $x$) (payoff P2 $y$) (payoff P3 $z$)})} assigns payoffs of $x$, $y$, and $z$, to the first, second, and third player, respectively.

\section{Proof of Equivalence} \label{Sec:ProofEquivalence}

Based on the strategy for constructing a Ludii game $\mathcal{G}^L$ for any extensive-form game $\mathcal{G}$ as described above, we present \reftheorem{Theorem:Equivalence}, which may intuitively be understood as stating that it is possible to model any arbitrary finite extensive-form game in L-GDL.

\begin{theorem} \label{Theorem:Equivalence}
Under Assumptions \ref{Assumption:InitialGameState} and \ref{Assumption:Player1First}, for any arbitrary extensive-form game $\mathcal{G}$, a corresponding Ludii game $\mathcal{G}^L$ constructed as described in Subsections~\ref{Subsec:DefiningPlayers}--\ref{Subsec:DefiningEndRules}, is equivalent to $\mathcal{G}$ in the sense that the following criteria are satisfied:
\begin{enumerate}
    \item \label{Criterion:ValidGameDescription} The game description of $\mathcal{G}^L$ is a valid game description according to the specification of L-GDL's grammar \cite{Browne_2020_LLR}.
    \item There exists a one-to-one correspondence between trajectories from the root node to any possible leaf node in the game tree of $\mathcal{G}$, and trajectories of play that are possible from the initial game state in $\mathcal{G}^L$. More concretely, this means that:
    \begin{enumerate}
        \item \label{Criterion:EquivalentStatesExist} For every node $s$ that is reachable from the root node $s_0$, including $s_0$ itself, there exists an equivalent game state in $\mathcal{G}^L$ that is also reachable in the same number of transitions from the initial game state of $\mathcal{G}^L$.
        \item \label{Criterion:CorrectMover} For every node $s$ in the game tree of $\mathcal{G}^L$ where $\iota(s) \neq \eta$ (i.e., any node that is not a chance node), the equivalent state in $\mathcal{G}^L$ also has $\iota(s)$ as the player to move.
        \item \label{Criterion:BranchingFactor} For every node $s$ in the game tree of $\mathcal{G}^L$ where $\iota(s) \neq \eta$, if there are $n$ branches to $n$ successors, there are also $n$ legal moves in the equivalent state in $\mathcal{G}^L$.
        \item \label{Criterion:ChanceNodes} For any chance node $s$ that is reachable from the root node $s_0$ of the game tree of $\mathcal{G}$, for every possible $s'$ that has a probability $\mathcal{D}(s, s') > 0$ of being the successor of $s$, there is also a probability $\mathcal{D}(s, s')$ that a transition to the equivalent state of $s'$ in $\mathcal{G}^L$ is the only legal transition in any arbitrary trajectory that reaches the equivalent of $s$ in $\mathcal{G}^L$.
        \item \label{Criterion:Payoffs} For any terminal node $s \in \mathcal{S}_{ter}$ in the game tree of $\mathcal{G}$, the equivalent state in $\mathcal{G}^L$ is also terminal, and assigns the same vector of payoffs $\mathcal{U}(s)$.
    \end{enumerate}
    \item \label{Criterion:InformationSets} Any player $1 \leq p \leq k$ playing the Ludii game $\mathcal{G}^L$ cannot distinguish between any pair of states that are the equivalents of two distinct nodes $s$, $s'$ if and only if they share the same information set $\mathcal{I}(p, s) = \mathcal{I}(p, s')$.
\end{enumerate}
\end{theorem}
These criteria are similar to those used for the proof of universality for S-GDL \cite{Thielscher_2011_Universal}.

\begin{proof}
By construction, the game description as detailed in subsections \ref{Subsec:DefiningPlayers} to \ref{Subsec:DefiningEndRules} is a valid L-GDL description. As of the public \texttt{v1.1.17} release of Ludii---which first introduced the \texttt{(random $\dots$)} and \texttt{(payoffs $\dots$)} ludemes---all of the ludemes used are supported. This satisfies criterion \ref{Criterion:ValidGameDescription}.

The start rules of $\mathcal{G}^L$ (see \reffigure{Fig:ProofStartRules}) ensure that, in the initial game state, a piece of type \texttt{Marker0} is placed on vertex $0$, and not on any other position. All moves that can possibly be generated are of the form depicted in \reffigure{Fig:ProofMoveRule}, which can only affect the positions of \texttt{Marker0} pieces through its \texttt{(fromTo (from $i$) (to $j$))} rule, which moves whichever piece is at vertex $i$ to vertex $j$. This means that the number of \texttt{Marker0} pieces cannot change; there must always be one, and only its position can change due to \texttt{(fromTo (from $i$) (to $j$))} rules. For any particular value of $i$, such a rule is only used in situations that satisfy the \texttt{(= (where "Marker" Neutral) i)} condition, i.e. only if vertex $i$ currently contains the sole \texttt{Marker0} piece. For any pair of values $i$ and $j$, if there is a branch from $s_i$ to $s_j$ in the game tree of $\mathcal{G}$, it is also possible for there to be a legal move that moves the \texttt{Marker0} piece from vertex $i$ to vertex $j$ in $\mathcal{G}^L$; such a move is either legal for sure if $s_i$ is not a chance node, or legal with probability $\mathcal{D}(s_i, s_j)$ if $s_i$ is a chance node. This satisfies criterion \ref{Criterion:EquivalentStatesExist}; the equivalent state of a node $s_i$ can always be identified as the one that has the \texttt{Marker0} piece on vertex $i$.

The move rules as described in \reffigure{Fig:ProofMoveRule} have, by construction, been set up to ensure that the next player to move is set to $\iota(s_j)$---or any arbitrary integer in $[1, k]$ if $\iota(s_j) = \eta$---whenever a move is made that moves the \texttt{Marker0} piece to vertex $j$---which means that the equivalent state of a node $s_j$ is reached. This ensures that criterion \ref{Criterion:CorrectMover} is satisfied for every node except for the root node $s_0$. \refassumption{Assumption:Player1First} ensures that the criterion is also satisfied for $s_0$.

By construction, as described in \refsubsection{Subsec:DefiningPlayRules}, for every node $s_i$ that is not a chance node, the equivalent state in $\mathcal{G}^L$ has its move rules defined by an \texttt{(or \{ $\dots$ \})} rule that wraps around $n$ different rules, each of which generates exactly $1$ legal move, such that $n$ is the number of successors of $s_i$ in the game tree of $\mathcal{G}$. This satisfies criterion \ref{Criterion:BranchingFactor}. Similarly, the correct number of moves with correct probabilities $\mathcal{D}(s, s')$ as required by \ref{Criterion:ChanceNodes} are explicitly defined as described in \refsubsection{Subsec:DefiningPlayRules}.

The end rules as described in \refsubsection{Subsec:DefiningEndRules} explicitly detect any game state in the Ludii game $\mathcal{G}^L$ that is the equivalent of a terminal node $s \in \mathcal{S}_{ter}$ of the extensive-form game $\mathcal{G}$, and explicitly assign the corresponding payoffs vector $\mathcal{U}(s)$. This satisfies criterion \ref{Criterion:Payoffs}.

By \refassumption{Assumption:InitialGameState}, there is only a single initial game state, and every player is aware of that. Therefore, every player's information set for the root node contains only the root node; $\forall_{p \in \{1, \dots, k\}} \mathcal{I}(p, s_0) = \{s_0\}$. This is reflected by the start rules described in \refsubsection{Subsec:DefiningStartRules} which, for every player $p$, place a marker for that player---and only visible to that player---in the subgraph used to represent the state space of $\mathcal{G}$ for that player. Every move that can be applied in any trajectory is of the form illustrated by \reffigure{Fig:ProofMoveRule}, which ensures that:
\begin{enumerate}
    \item Every player $p$ can only ever observe markers on vertices of ``its own'' subgraph.
    \item Let $j$ denote the vertex that contains the neutral marker---hidden from \emph{all} players---in the first subgraph. For every player $p$, within that player's ``own'' subgraph, there is always a marker on every vertex that represents any of the nodes in the information set $\mathcal{I}(p, s_j)$ for that player in that state.
\end{enumerate}
This means that, for any pair of nodes that is in the same information set for a player in the game tree of $\mathcal{G}$, the pair of equivalent game states in $\mathcal{G}^L$ are also indistinguishable from each other from that player's perspective (due to the arrangement of markers on vertices visible to that player being identical). Note that the move rules as described in \reffigure{Fig:ProofMoveRule} were deliberately set up such that players always select vertex $n$ to pick the $n^{th}$ move in a list of legal moves, irrespective of which vertices are subsequently affected by that move. It might have been more intuitive to directly select the vertex corresponding to the node in the extensive-form game tree to transition into, but this could reveal additional information that the player should not have access to. With this, criterion \ref{Criterion:InformationSets} is also satisfied and the proof is complete.
\end{proof}

\section{Discussion} \label{Sec:Discussion}

The main topic of this paper, with \reftheorem{Theorem:Equivalence} and its proof, is to prove that L-GDL is sufficiently expressive to model the equivalent of any arbitrary finite extensive-form game. A related question of potential interest is the converse of \reftheorem{Theorem:Equivalence}: \textit{is any game that can be modelled in L-GDL equivalent to a finite extensive-form game?} This question can be answered in the negative with counterexamples. For example, the game of Mu Torere (which has been implemented in Ludii) is known to go on indefinitely under perfect play \cite{Ascher_1987_MuTorere}, which makes it an example of a game with an infinitely-sized game tree that can be implemented in L-GDL. Hence, while L-GDL is sufficiently expressive to model any game of the class considered in \reftheorem{Theorem:Equivalence}, it is not restricted to that class.

\section{Conclusion} \label{Sec:Conclusion}

Ludii's game description language (L-GDL) has primarily been designed to be easy to use for game designers, with a focus on facilitating the design of board games and similar abstract games. In practice, its count of over 1000 distinct game descriptions\footnote{\url{https://ludii.games/library.php}} (which far exceeds the game counts in official repositories of many other systems with GDLs, such as S-GDL \cite{Genesereth_2005_GGP,Love_2008_GDL}, RBG \cite{Kowalski_2019_Regular}, and GVGAI \cite{Perez_2019_GVGAI}) has already demonstrated its flexibility and generality. In this paper, we have also proven its generality from a theoretical angle, demonstrating that it is possible to write an equivalent game in L-GDL for any arbitrary finite extensive-form game \cite{Rasmusen_2007_Games}. Two assumptions (Assumptions~\ref{Assumption:InitialGameState} and~\ref{Assumption:Player1First}) on the structure of extensive-form games were made to simplify the proof, but both assumptions are without loss of generality. Simultaneous-move games were not considered explicitly, but are implicitly also covered by the proof due to the possibility of modelling any simultaneous-move game as a sequential one with hidden information \cite{Watson_2013_Strategy}. This provides a significant extension of an earlier proof \cite{Piette_2020_Ludii} by including stochastic and imperfect-information games (and, implicitly, also simultaneous-move games), and means that the expressiveness of L-GDL matches that proven by Thielscher \cite{Thielscher_2011_Universal} for S-GDL.

This result suggests that we can opt to use Ludii over S-GDL in GGP research for some of its other advantages, such as computational efficiency \cite{Piette_2020_Ludii} and ease of use, without a loss in expressiveness. While the somewhat convoluted way of defining games used for the theoretical proof is unlikely to be an efficient way of implementing many ``real'' games in practice, it may be a fruitful starting point for designing synthetic game trees for targeted research into the relations between certain game tree characteristics and the effectiveness of different algorithms \cite{Ramanujan_2010_Understanding,Ramanujan_2011_Synthetic} within the same framework and API (Ludii) that also supports many real games.

In this paper, we primarily focused on the expressiveness of the L-GDL language. This is one of the primary types of properties that is typically considered of importance for GDLs \cite{Thielscher_2011_Universal,Kowalski_2019_Regular}. However, there are also other interesting theoretical properties of GDLs and game descriptions, which may be further explored (for the case of L-GDL) in future work. For example, in future work it would be interesting to examine whether or not L-GDL is Turing complete, and what the complexity is of deciding whether or not any given L-GDL description satisfies certain properties such as playability or well-formedness \cite{Saffidine_2014_TuringComplete}.

\section*{Acknowledgment} 
This research is funded by the European Research Council as part of the Digital Ludeme Project (ERC Consolidator Grant \#771292).

\bibliographystyle{IEEEtran}
\bibliography{dlp-biblio-1}

\end{document}